# Real-time policy generation and its application to robot grasping

Masoud Baghbahari, Aman Behal

Department of Electrical Engineering and NanoScience Technology Center
University of Central Florida
Orlando, FL32816 USA

bahari@knights.ucf.edu, abehal@ucf.edu

## ABSTRACT
Real time applications such as robotic require real time actions based on the immediate available data. Machine learning and artificial intelligence rely on high volume of training informative data set to propose a comprehensive and useful model for later real time action. Our goal in this paper is to provide a solution for robot grasping as a real time application without the time and memory consuming pertaining phase. Grasping as one of the most important ability of human being is defined as a suitable configuration which depends on the perceived information from the object. For human being, the best results obtain when one incorporates the vision data such as the extracted edges and shape from the object into grasping task. Nevertheless, in robotics, vision will not suite for every situation. Another possibility to grasping is using the object shape information from its vicinity. Based on these Haptic information, similar to human being, one can propose different approaches to grasping which are called grasping policies. In this work, we are trying to introduce a real time policy which aims at keeping contact with the object during movement and alignment on it. First we state problem by system dynamic equation incorporated by the object constraint surface into dynamic equation. In next step, the suggested policy to accomplish the task in real time based on the available sensor information will be presented. The effectiveness of proposed approach will be evaluated by demonstration results.

## Keywords
Real time, Policy generation, Dynamic equation, Grasping.

## 1. INTRODUCTION
In the era of mining and interpreting data, machine learning and artificial intelligence have demonstrated their abilities in this regard. A lot of works and attempts reported in literatures specifically during recent years provide better insight about the efficacy of the mentioned fields [2]. In fact, the pursued goal behind all of these recent valuable researches is to present a simplified, comprehensive parametric or non-parametric representation for modeling, decoding and extracting the informative features underneath the provided data set. The obtained outcome will be utilized to predict an output corresponding to new input data [3].

An essential assumption which highly determines the quality of prediction is the informativeness of data set itself. A machine learning algorithm always works based on previous observed and known data as the main source to understand the underlying unknown structure. Having access to more volume and a rich data set certainly helps the algorithm to differentiate between various of future unknown input data and interpolate better to provide more accurate results [1].

When it comes to the realm of real time applications more considerations will be required. Real time applications are always facing with possible new different data at each instance of the time. The time-dependent nature of these data gives a dynamic behavior to the data set. For example, an autonomous vehicle moving in surrounding with other vehicles, pedestrians and under traffic restrictions needs to react immediately according to the received real time data from its sensors, camera or remote information. As another example consider a networks of connected sensor nodes transferring data between themselves or a master node. Different routes for this purpose are available between them and selection of best one needs a real time analyzing process [5]-[4]-[11].

A robot as an agent dealing with real time unknown environment and structure needs to adjust its behavior dynamically with such time and structure dependent changes. For instance, a robot aims at grasping an unknown object. The expected shape of real objects ranges from the simplest form to the most complex one. To grasp a target object a robot has to adapt the limb and end-effector configuration. The configuration depends on the object shape feature which is the edges of the object. For human being, the information of edges can be achieved by vision inspection easily. Several researches have been conducted using the same idea and ability by a camera in robotic [7] – [8]. Nonetheless, despite enhancement, vision algorithm in robotic domain still suffers from several weakness. High time computational and memory usage for a light, low specifications and remote control robot analyzing several other sensor information at the same time are the expense of advanced vision algorithm implementation [8].

Meanwhile, another feasible option is to precept the object through touching and sensing the surface. This Haptic strategy has been addressed in several references promising to perform better under the explained limitations in robotics. Fortunately, new manufactured robot has been equipped with sensors which gives the contact force torque information at each instant of the time. One can gather such kind of information from the object as the input data set to feed them to machine learning algorithms for structure identification, classification, modeling or shape reconstruction. Certainly previous similar diverse circumstances learning are needed to deal with novel and unknown objects. Moreover, the trained model which can be very complex should be stored in robot internal memory for real time implementation [8].

In this paper we will propose the solution for such machine learning shortages by directly implementing a policy for grasping by the robot. In other words, the idea would be eliminating the learning and exploring process as the intermediate level, proposing an immediate policy in real time to accomplish the grasping task as the robot keeps touching the object. The paper will be organized as different sections. Section 2 describes the system dynamic behavior and object force torque generating model, in section 3 the unified real time policy will be presented. The demonstrations of work are provided in section 4 and section 5 summarizes the paper.

## 2. SYSTEM DESCRIPTION

In this section, the required mathematical relationship of system in real time and the model of interaction surface are investigated. The provided mathematical relationships are useful for next sections algorithm.

### 2.1 System Real Time Dynamic Equation

To derive a real time policy, a time-dependent model of system is required. Such model is so called dynamic equation in robotic expressing the relationship between target variables and the input commanded signal. The target variables are either joint velocities $\dot{q}$ or the end-effector velocity $\dot{x}$ respect to a specific frame. The end-effector position and orientation is a twist vector of size of $(6 \times 1)$ denoted by $x \in \mathbb{R}^6$. The time derivative of these variables so called linear and angular velocity is related to joint velocities by Jacobian matrix $J(q)$ :

$$J(q)\dot{q} = \dot{x} \quad (1)$$

On the other hand, grasping is a kind of robot environment interaction. As a consequence, it is essential to evaluate the effect of this interaction on the robot dynamic equation. Keep in mind this fact, the dynamic equation of robot in joint space is described by:

$$M(q)\ddot{q} + C(q,\dot{q}) + g(q) = \tau - J^T F \quad (2)$$

$\tau$ is the joint torques vector, $F$ representing the interaction force-moment between the robot end-effector and the object. The inserted force and moment by end-effector to object form the components of wrench $F$:

$$F = \begin{bmatrix} f_e \\ m_e \end{bmatrix} \quad (3)$$

Inertia matrix $M(q)$ which is dependent on the joint angles is a bounded positive definite matrix, the Centrifugal and Coriolis forces $C\dot{q}$ and $g(q)$ gravitational forces are both bounded as well. The skew-symmetricity is one of the most important feature of robot [12].

Depend on the number of joints, redundancy in robot kinematics is possible. An immediate consequence of redundancy is smaller dimensionality of Cartesian space than joint space. Assuming that $J(q)$ is a full row rank matrix, further decomposition of joint velocity vector achieves by [12]:

$$\dot{q} = J^+ J\dot{q} + J^- \dot{q} \quad (4)$$

Where $J^+ = J^T(JJ^T)^{-1}$ and $J^- = I - (J^+J)$.

To have more control over the end-effector situation over time, it is more convenient to consider the Cartesian space dynamic equation. To do that, we can transform the joint space dynamic equation to Cartesian space by taking derivative of (1) and express the joint acceleration according to end-effector acceleration as follow:

$$\ddot{x} = J\ddot{q} + \dot{J}\dot{q} \quad (5)$$

$$\ddot{q} = J^+(\ddot{x} - \dot{J}\dot{q}) + J^-\ddot{q} \quad (6)$$

Replacing into (2), the final dynamic equation would be:

$$MJ^+ \ddot{x} + MJ^-\ddot{q} + g + d(t) = \tau - J^T F \quad (7)$$

In which the $d(t) = -MJ^+\dot{J}\dot{q} + C\dot{q}$ is a very small magnitude term and will be considered as disturbance. To create a better relationship between the end-effector and interacting force moment, following torque signal $\tau$ is presented as the first step of policy generation:

$$\tau = MJ^+(\ddot{x}_d - M_d^{-1}(B_d(\dot{x} - \dot{x}_d) - (F - F_d))) + J^T F \quad (8)$$

This torque signal simply imply the cancellation of force-moment effect via jacobian matrix on the joint torque and brings the desired force regulation inside a decoupled dynamic equation. By this way the force-moment effect in each direction can be handled much more easily.

For grasping the target end-effector velocity variable respect to end-effector itself provides more freedom to manipulate and guide the end-effector during alignment and movement on the object. To achieve this goal, the acceleration signal designs base on the relationship between the end-effector velocity in base frame and the end-effector velocity in end-effector frame:

$$\dot{x} = R_e \dot{x}_e \quad (9)$$

$$\ddot{x} = \bar{R}_e \ddot{x}_e^e + \dot{\bar{R}}_e \dot{x}_e^e \quad (10)$$

In which we have:

$$\bar{R}_e = \begin{bmatrix} R_e & 0 \\ 0 & R_e \end{bmatrix} \quad (11)$$

Now it is enough to have a choice for acceleration as:

$$a = \bar{R}_e a^e + \dot{\bar{R}}_e \dot{x}_e^e \quad (12)$$

Such selection simply provides:

$$\ddot{x}_e^e = a^e \qquad (13)$$

$a$ is the acceleration referred to the end-effector frame.

This equation simply means that to change the system behavior at each instant of the time, suitable acceleration signal as a good policy has to be generated. The best policy obtains when we incorporate feedback from the action into the policy. The reason behind that is because any degradation from the desired policy has to be compensated in a closed loop manner. To reach to this goal, we propose following policy for acceleration:

$$a^e = M_d^{-1}\,(\dot{v}_d^e + B_d(v_e^e - v_{de}^e) - B_f(F_e - F_{de})) \qquad (14)$$

Based on this expression, the robot configuration can alter according to any terms in right hand side of equation. To put in to perspective, the result from inserting this policy into robot dynamic equation (7), finally one can reach to the following dynamic equation known as error dynamic:

$$M_d\,(\dot{v}_e^e - \dot{v}_{de}^e) + B_d(v_e - v_{de}^e) = B_f(F_e^e) \qquad (15)$$

Because the coefficients of this dynamic error are positive, starting from any initial conditions for the velocity, the error signal for each term will be vanished over time. Using appropriate coefficients for $M_d, B_d$ the rate of convergence to zero can be adjusted easily.

As it is also shown in experimental results section, some recent advanced manufactured robots have provided a velocity mode control in which the robot can be easily commanded in desired direction of movement by a velocity command. Such mode of control automatically will compensate the effect of external force torque on the joints, a direct dynamic equation of the robot velocity would be accessible without need of inertia and Coriolis terms parameters involved in mathematical equations.

## 2.2 Interacting Force Moment Modeling

To introduce a model for interacting force-moment between the object and end-effector, similar to [12] we assume that the object 3D surface geometry can be stated as a 3-dimensional constraint manifold by:

$$\phi(X,\theta) = 0 \qquad (16)$$

During the contact with the object the end-effector Cartesian position satisfies the above equation. The $\theta$ is a vector including the object-related parameters and determines the shape of the object or constrained surface.

Using partial derivative and chain role taking time derivative of (16) yields following equation:

$$J(X,\theta)\dot{X} = 0 \qquad (17)$$

In which $J(X,\theta)$ is the so called constraint Jacobian matrix defined as:

$$J(X,\theta) = \frac{\partial \phi(X,\theta)}{\partial X} \qquad (18)$$

This relationship indicates that the velocity of end-effector movement on the surface is always perpendicular to the surface gradient vector. Actually, the velocity of end-effector belongs to the null space of constraint jacobian matrix. This matrix is very useful in modeling of moment inserted from the surface to the end-effector. The moment generates under the end-effector attached sensor from the misalignment situation between the sensor plate with the surface. Having access to the surface gradient vector and the end-effector direction, one can determine the resulting moment. The interaction force between the end-effector and the surface directly depends on the amount of elasticity of the contacted object, Fig. 1.

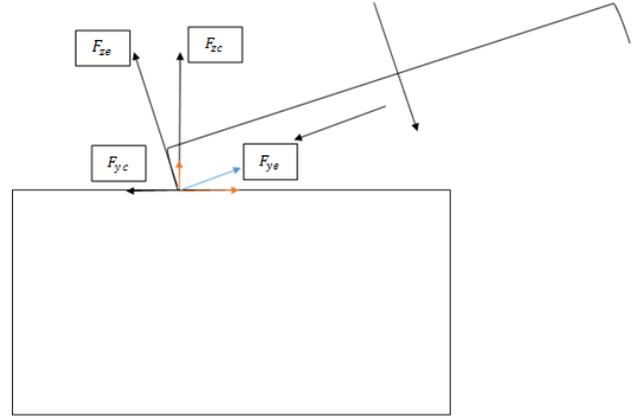

**Figure 1. Decomposition of contact force during grasping**

Assuming the elasticity of $K$, interaction force is also related to the amount of surface deformation from the equilibrium point of contact $X_0$:

$$F_e = K|X - X_0| \qquad (19)$$

The force is always in the end-effector z direction or end-effector approaching direction.

To formulate the moment, let to assume that the wrist has a circle shape with radios equal $r$. The resulting moment sensing under the sensor plate would be:

$$m_e = r(z_e \times J_e f_e) \qquad (20)$$

This relationship indicates that the cross product of end-effector approaching direction and surface normal vector, is a measure of misalignment between the end-effector and the object surface.

Where $z_e$ is the unit vector of end-effector approaching direction is expressed in end-effector frame and $J_e$ is the surface gradient vector $f_e$ is the force vector capturing the end-effector z direction inserted force by (20) defined as follow:

$$f_e = \begin{bmatrix} 0 \\ 0 \\ F_e \end{bmatrix} \quad (21)$$

In next section, using the proposed model for force and moment modeling, we aim to generate required force moment term inserted in right hand side of (15) to guide the end-effector to move on the object surface. This movement finally reaches the end-effector to a suitable configuration for grasping the target object.

## 3. UNIFIED REAL TIME POLICY GENERATION

Grasping is a task of combination of movement and alignment around the object. Depend on the rate of mixture of movement and alignment, the object will be grasped first either by movement or alignment. Naturally this process can be done sequentially or continuously. Sequentially means first completely alignment with the object and then move on the object and vise-versa. As these two distinguishable phases be accomplished at the same time the process is called continuously. Our goal is to provide a smooth combined version of mixture of both of them. All these explanation reveals this fact that grasping task is required a kind of policy definition for the task at each instant of time. In other words, a real time on line version of policy generation is required.

Aiming to this purpose, the following definition for $F_e^e$ in (15) is presented:

$$F_e^e = \begin{bmatrix} 0 \\ (1 - \alpha(\tanh(F_{ye})))(F_{ye} - F_{dye}) \\ F_{ze} - F_{dze} \\ \tau_{xe} \\ \tau_{ye} \\ 0 \end{bmatrix} \quad (22)$$

All the terms inside the vector is the sensor provided information. The role of $(1 - \alpha(\tanh(F_{ye})))$ is crucial. As long as the force in movement direction or y direction is near zero, the whole coefficient vanishing and the commanded force in y direction guides the end-effector on the surface. Also, if it is far from zero, it is an indicator of misalignment with the object. In such case, alignment has more priority over movement. So the whole term is around zero which means ineffective commanded force in y direction.

More precisely, the term in end-effector y direction can move the end-effector as long as the force in opposite direction be less than predefined desired force. This situation certainly cannot happen when the end-effector is not aligned with the surface. Here is where the moment on the end-effector due to misalignment isn't zero. As a consequence of commanding the robot with this measured moment from the sensor, the robot starts to become aligned with the object as much as it can to reduce the level of opposite force and create freedom for end-effector to move on the object in desired direction.

## 4. Experimental Results

To demonstrate the proposed real time policy, we use the Mico Kinova robot. It is a six degree of freedom light assistive robot supported by ROS package. The robot has velocity mode control in which one can command the robot end-effector with desired velocity. Based on (9) using the rotation matrix between the end-effector and base frame, one can easily compute the end-effector velocity in end-effector frame and also send the desired velocity in end-effector frame to the robot. To identify the dynamic equation governs the system, we can command the system by just step function in each direction of end-effector. The error between the command and the end-effector velocity is in Figure.2. Since the data received form the robot is very noisy, we just pass them through a low pass filter.

The target object for grasping is a bottle of water. We select $\alpha$ the coefficient of $tanh$ function in (20) equal zero. In such case, the grasping task will be done purely as an alignment before complete grasping. As the uploaded video in (https://www.youtube.com/watch?v=Y19Zw3-3VDY) confirms, the grasping by proposed policy is done successfully. The Figure.3 shows the profile of end-effector force and moment during making contact with the object. The force in approaching direction is set on a constant value while the end-effector has contact with the object during grasping process. It is worth noting that the force for x and y direction are due to this fact that the gripper fingers have no equal contact with the object. The alignment is done by the information provided by the torque around end-effector y axis denoted by $\tau_y$. All the force torque sensor data converge to a steady state after a period of alignment. During this time the torque results from misalignment is bigger that it's steady state situation.

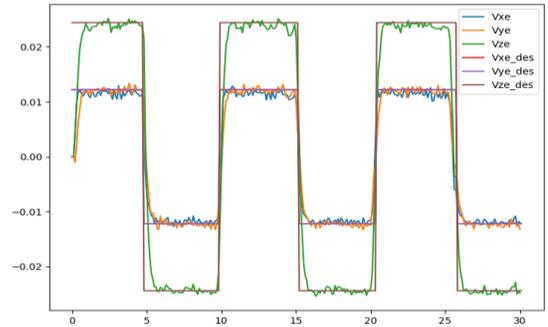

**Figure 2. End-Effector Frame Velocity and Desired Values**

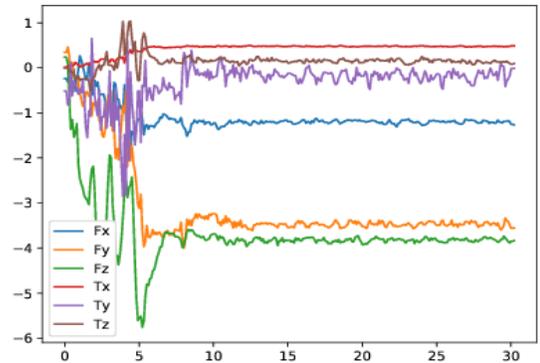

**Figure 3. The force moment profile during alignment and grasping the object**

## 5. CONCLUSIONS

In this paper we proposed a real time policy generation algorithm based on the real time data. The approach is smooth in implementation. Robot grasping as a good example demonstrates the performance of suggested solution. The provided figures of work indicate that the robot can align itself with object during grasping process.

## 6. REFERENCES

[1] Robert, C. (2014). Machine learning, a probabilistic perspective.
[2] N. H. Bidoki, M. B. Baghdadabad, "A glance at structural classifiers and their applications," 2nd National Conference on New Approaches in Computer Engineering Islamic Azad University, Roudsar-Branch.
[3] Alpaydin, Ethem, "Introduction to machine learning," MIT press, 2014.
[4] N. H. Bidoki, M. B. Baghdadabad, G. R. Sukthankar, and D. Turgut,"Joint Value of Information and Energy Aware Sleep Scheduling in Wireless Sensor Networks: A Linear Programming Approach," In Accepted to appear in IEEE ICC'18, May 2018.
[5] Nishihara, Robert, et al, "Real-time machine learning: The missing pieces.," Proceedings of the 16th Workshop on Hot Topics in Operating Systems. ACM, 2017.
[6] N. H. Bidoki and M. D. T. Fooladi, "Linear programming-based model for joint routing and sleep scheduling in data-centric wireless sensor networks," Information and Knowledge Technology (IKT), 2014.
[7] Levine, Sergey, et al. "Learning hand-eye coordination for robotic grasping with large-scale data collection," International Symposium on Experimental Robotics. Springer, Cham, 2016.
[8] Bohg, Jeannette, et al. "Data-driven grasp synthesisa survey " IEEE Transactions on Robotics 30.2 (2014): 289-309.
[9] A. Mayle, N. H. Bidoki, S. S. Bacanli, L. Blni, and D. Turgut, "Investigating the Value of Privacy within the Internet of Things," In Proceedings of IEEE GLOBECOM'17, December 2017.
[10] D. Turgut, L. Massi, N. H. Bidoki, and S. S. Bacanli "Multidisciplinary Undergraduate Research Experience in the Internet of Things: Student Outcomes, Faculty Perceptions, and Lessons Learned," 2017 ASEE Annual Conference Exposition Conference, June 2017. National Conference, 2016.
[11] M. Baghbahari, N. Hajiakhoond "Evaluation of estimation approaches on the quality and robustness of collision warning systems," IEEE SoutheastCon, April 2018.
[12] Namvar, Mehrzad, and Farhad Aghili "Adaptive force-motion control of coordinated robots interacting with geometrically unknown environments, " IEEE Transactions on Robotics 21.4 (2005): 678-694.